\documentclass[10pt,twocolumn,letterpaper]{article}

\usepackage{iccv}              
\usepackage{multirow}
\usepackage{colortbl}
\definecolor{iccvblue}{rgb}{0.21,0.49,0.74}
\usepackage[pagebackref,breaklinks,colorlinks,allcolors=iccvblue]{hyperref}

\def\paperID{2146} 
\def\confName{ICCV}
\def\confYear{2025}

\title{Boosting Multi-View Indoor 3D Object Detection via \\ Adaptive 3D Volume Construction}

\author{
	Runmin Zhang$^{1}$\quad
	Zhu Yu$^{1}$\thanks{Corresponding authors.}\quad
	Si-Yuan Cao$^{2,3}$\footnotemark[1]\quad
	Lingyu Zhu$^{4}$\\
	Guangyi Zhang$^{1}$\quad
	Xiaokai Bai$^{1}$\quad
	Hui-Liang Shen$^{1}$ \\
	{\small{$^{1}$College of Information Science and Electronic Engineering, Zhejiang University}} \\
	{\small{$^{2}$Ningbo Global Innovation Center, Zhejiang University}}\quad
	{\small{$^{3}$NingboTech University}}\quad
	{\small{$^{4}$City University of Hong Kong}} \\
	{\tt\small \{runmin\_zhang, yu\_zhu, cao\_siyuan\}@zju.edu.cn, lingyzhu-c@my.cityu.edu.hk,} \\
	{\tt\small \{zhangguangyi, shawnnnkb, shenhl\}@zju.edu.cn}
}
	
\begin{document}
\maketitle

\begin{abstract}
This work presents SGCDet, a novel multi-view indoor 3D object detection framework based on adaptive 3D volume construction. Unlike previous approaches that restrict the receptive field of voxels to fixed locations on images, we introduce a geometry and context aware aggregation module to integrate geometric and contextual information within adaptive regions in each image and dynamically adjust the contributions from different views, enhancing the representation capability of voxel features. Furthermore, we propose a sparse volume construction strategy that adaptively identifies and selects voxels with high occupancy probabilities for feature refinement, minimizing redundant computation in free space. Benefiting from the above designs, our framework achieves effective and efficient volume construction in an adaptive way. Better still, our network can be supervised using only 3D bounding boxes, eliminating the dependence on ground-truth scene geometry. Experimental results demonstrate that SGCDet achieves state-of-the-art performance on the ScanNet, ScanNet200 and ARKitScenes datasets. The source code is available at \url{https://github.com/RM-Zhang/SGCDet}.
\end{abstract}

\section{Introduction}
\label{sec:intro}
Indoor 3D object detection is a fundamental 3D perception task, with broad applications in embodied AI, AR/VR, and robotics. Leveraging precise scene geometry as input, point cloud-based 3D object detectors~\cite{votenet_iccv19, h3dnet_eccv20, fcaf3d_eccv22, cagroup3d_nips22, unidet3d_arxiv24, cascadev_arxiv24} have achieved impressive performance. However, capturing accurate scene geometry typically requires high-cost 3D sensors. Recently, there has been a shift towards using multi-view posed images for 3D object detection.

\begin{figure}[t]
	\centering
	\includegraphics[scale=0.65] {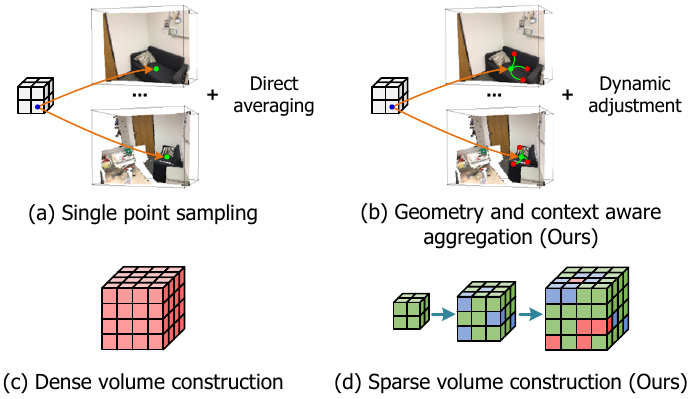}
	\vspace{-2mm}
	\caption{Comparison of feature lifting and volume construction strategies between previous approaches and our SGCDet. (a) The single point sampling strategy used in previous approaches restricts the receptive field of voxels to a limited region, neglecting contextual information across multiple views. (b) Our geometry and context aware aggregation adaptively integrates geometric and contextual features within deformable regions across different views, enhancing the representational capability of voxel features. (c) Previous approaches construct high-resolution, dense 3D volumes without considering the inherent sparsity of 3D scenes, leading to unnecessary computational overhead. (d) Our sparse volume construction adaptively refines voxels that are likely to contain objects, reducing redundant computational cost in free space.}
	\label{fig:head}
	\vspace{-6mm}
\end{figure}

To bridge the gap between 2D images and 3D representations, the pioneering work ImVoxelNet~\cite{imvoxelnet_wacv22} lifts 2D features along the overall ray. Each voxel then adopts the averaged results across multiple views as its features. However, this approach uses the same weights for features derived from different images, leading to a coarse and error-prone 3D voxel representation. Although the following works~\cite{imgeonet_iccv23, nerfdet_iccv23} introduce an opacity probability to suppress voxel features in free space through post-processing, they still fail to address the occlusion issue during the 2D-to-3D projection process. More recent approaches~\cite{cnrma_cvpr24, mvsdet_nips24} introduce explicit geometry constrains to assist the feature lifting. Nevertheless, the final performance of these approaches heavily depends on the accuracy of the estimated geometric information, either complicating the training pipeline or significantly increasing computational cost.

As analyzed above, previous approaches primarily enhance the quality of 3D voxel representations from a geometric perspective, overlooking the valuable contextual information of images. The sampling locations on 2D feature maps are constrained to fixed positions determined by the predefined voxel centers and camera poses. This single point sampling strategy limits the receptive field of voxels to a small region, restricting their ability to perceive visual information. In addition, this strategy further amplifies the dependency on accurate geometric information~\cite{cnrma_cvpr24, mvsdet_nips24}, as illustrated in Fig.~\ref{fig:head}(a).

To address these issues, we propose a \textbf{geometry and context aware aggregation} module to adaptively lift the 2D features. Instead of simply performing a weighted average of the sampled features across multi-view images, we take the sampled features as queries to aggregate relevant geometric and contextual features within a deformable region. Furthermore, we introduce a multi-view attention mechanism to dynamically adjust the contributions from different views, enhancing the representation capabilities of the transformed 3D volumes, as illustrated in Fig.~\ref{fig:head}(b).

On the other hand, previous approaches generally construct high-resolution, dense 3D volumes, as shown in Fig.~\ref{fig:head}(c). This dense representation fails to account for the inherent sparsity of 3D scenes, leading to unnecessary computational overhead. To address this issue, we propose a \textbf{sparse volume construction} strategy that constructs 3D volumes in an adaptive manner, as illustrated in Fig.~\ref{fig:head}(d). Specifically, we employ an occupancy prediction module to identify voxels likely to contain objects for refinement, thereby reducing redundant computations in free space. A critical aspect of this strategy is the supervision of occupancy prediction. While a straightforward solution is to directly use ground-truth geometry for supervision~\cite{imgeonet_iccv23, cnrma_cvpr24}, it is infeasible when such data is unavailable~\cite{mvsdet_nips24, objectron_cvpr21}. To eliminate reliance on ground-truth geometry, we leverage 3D bounding boxes to generate pseudo labels for occupancy, achieving flexible network supervision.

By combining the \textbf{S}parse volume construction and the \textbf{G}eometry and \textbf{C}ontext aware aggregation, we propose a novel framework for multi-view indoor 3D object \textbf{D}etection, named \textbf{SGCDet}. Thanks to above designs, SGCDet performs effective and efficient 3D volume construction. We evaluate our SGCDet on the ScanNet~\cite{scannet_cvpr17}, ScanNet200~\cite{scannet200_eccv22}, and ARKitScenes~\cite{arkitscenes} datasets. SGCDet achieves state-of-the-art performance among approaches that do not rely on ground-truth geometry for supervision. Compared to the previous state-of-the-art approach MVSDet~\cite{mvsdet_nips24}, SGCDet significantly improves mAP@0.5 by 3.9 on ScanNet, while reducing training memory, training time, inference memory, and inference time by 42.9\%, 47.2\%, 50\%, and 40.8\%, respectively. Remarkably, SGCDet also surpasses some approaches that use ground-truth geometry during training.

Our contributions are summarized as follows:
\begin{itemize}
	\item We propose the geometry and context aware aggregation module to enhance feature lifting. It enables each voxel to adaptively aggregate geometric and contextual features within a deformable region, and dynamically adjusts feature contributions across different views.
	
	\item We introduce the sparse volume construction strategy, which adaptively refines voxels likely to contain objects, reducing computations in free space. Notably, the overall network can be supervised using only 3D bounding boxes, eliminating the need for ground-truth geometry.
	
	\item Extensive experiments demonstrate that SGCDet outperforms the previous state-of-the-art approach by a large margin, while significantly reducing computational overhead. These results validate both the effectiveness and efficiency of SGCDet.
\end{itemize}

\section{Related Works}
\label{sec:related}

\begin{figure*}[t]
	\centering
	\includegraphics[scale=0.85] {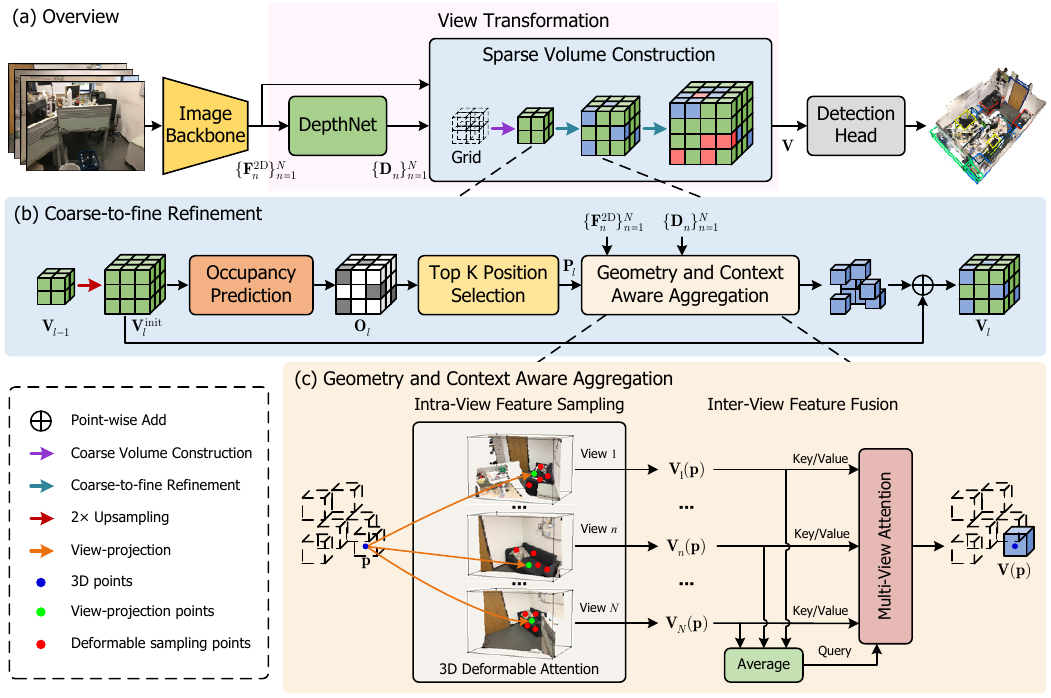}
	\vspace{-2mm}
	\caption{Schematics and detailed architectures of SGCDet. (a) Overview of SGCDet, which consists of an image backbone to extract image features, a view transformation module to lift image features to 3D volumes, and a detection head to predict 3D bounding boxes. (b) Details of the coarse-to-fine refinement in our sparse volume construction strategy. (c) Details of our geometry and context aware aggregation module.}
	\vspace{-6mm}
	\label{fig:method}
\end{figure*}

\textbf{Image-based 3D Object Detection.} Image-based 3D object detection has gained significant attention due to its cost-effectiveness and fine-grained visual perception capabilities. Current approaches primarily focus on constructing 3D representations from input images, including bird's-eye-view (BEV)~\cite{bevdet_arxiv21, bevdepth_aaai23, bevformer_tpami24, dfa3d_iccv23} and voxel-based approaches~\cite{imvoxelnet_wacv22, imgeonet_iccv23, nerfdet_iccv23, mvsdet_nips24, cnrma_cvpr24, embodiedscan_cvpr24}. Given the variability in camera viewpoints and object distributions, voxel-based representations are better suited for indoor scenes. ImVoxelNet~\cite{imvoxelnet_wacv22} is the pioneer that introduces an end-to-end pipeline for multi-view indoor 3D object detection. It directly lifts 2D features along 3D rays without incorporating scene geometry, leading to ambiguities in volume features. Building on ImVoxelNet, ImGeoNet~\cite{imgeonet_iccv23} and NeRF-Det~\cite{nerfdet_iccv23} compute opacity probabilities for the 3D volume to suppress features in free space. NeRF-Det++~\cite{nerfdet++_tip25} and Go-N3RDet~\cite{gon3rdet_cvpr25} further enhance NeRF-Det through semantic and geometric constraints. However, these approaches treat opacity solely as a post-processing step, and fail to address occlusion issues during the feature lifting process. Alternatively, recent approaches~\cite{cnrma_cvpr24, mvsdet_nips24} explicitly estimate scene geometry to achieve occlusion-aware projection. CN-RMA~\cite{cnrma_cvpr24} combines a 3D reconstruction network with a point cloud-based 3D object detector, and uses a reconstructed TSDF to guide the feature lifting process. Nevertheless, it requires a time-consuming multi-stage training pipeline, and relies on ground-truth geometry for supervision. In contrast, MVSDet~\cite{mvsdet_nips24} leverages multi-view stereo to compute depth probabilities from input images, and applies 3D Gaussian Splatting~\cite{mvsplat_eccv24} for self-supervision. However, it still suffers from high computational costs.

\textbf{Sparse Design in 3D Vision.}
Inspired by DETR~\cite{detr_eccv20}, several 3D detection methods~\cite{detr3d_corl22, petr_eccv22, parq_iccv23, sparsebev_iccv23} employ a sparse set of object queries to enable 3D-to-2D interaction. However, due to the absence of explicit 3D representations, these methods typically suffer from slow convergence. Other occupancy prediction approaches reduce the number of voxel queries through depth-based query proposal initialization~\cite{voxformer_cvpr23, cgformer_nips24, voxdet_arxiv25}, multi-scale sparse reconstruction~\cite{octreeocc_nips24, sparseocc_eccv24}, or by reformulating the problem as a sparse set prediction~\cite{opus_nips24}. While these methods have made notable progress, they still rely on precise geometry for supervision, limiting their applicability in scenarios where ground-truth geometric information is unavailable.

\section{Method}
\label{sec:method}

\subsection{Overview}
Given $N$ posed images $\{\mathbf{I}_{n}\}_{n=1}^{N}$ as input, SGCDet aims to predict 3D bounding boxes of the scene. As illustrated in Fig.~\ref{fig:method}(a), the overall framework of SGCDet consists of three main components: an image backbone that extracts 2D features $\{\mathbf{F}_{n}^\mathrm{2D} \in \mathbb{R}^{H \times W \times C}\}_{n=1}^{N}$, a view transformation module that lifts these 2D features to 3D volumes $\mathbf{V} \in \mathbb{R}^{X \times Y \times Z \times C}$, and a detection head that predicts 3D bounding boxes. Here, $(H, W)$ and $(X, Y, Z)$ represent the spatial resolution of 2D features and 3D volumes, respectively. $C$ denotes the number of channels.

Our core design focuses on the view transformation module, which achieves adaptive 3D volume construction. Specifically, we adopt a simple yet effective \textbf{DepthNet} (Sec.~\ref{subsec:depthnet}) to estimate the depth distributions $\{\mathbf{D}_{n} \in \mathbb{R}^{H\times W\times D}\}_{n=1}^{N}$ for the input images, where $D$ is the number of depth bins. The depth distributions provide geometric information for the view transformation process. To address the inefficiency of dense volume construction, we propose the \textbf{sparse volume construction} (Sec.~\ref{subsec:asvr}) that adaptively builds the 3D volume in a coarse-to-fine manner. Within this process, we introduce the \textbf{geometry and context aware aggregation} (Sec.~\ref{subsec:gcap}), which ensures adaptive feature lifting by integrating geometric and contextual information within a flexible region.

\subsection{Sparse Volume Construction}
\label{subsec:asvr}

Given the dense 3D grid $\mathbf{G} \in \mathbb{R}^{X \times Y \times Z \times 3}$, previous approaches~\cite{imvoxelnet_wacv22, imgeonet_iccv23, nerfdet_iccv23, mvsdet_nips24} typically project each 3D voxel to 2D features for volume construction. However, since most voxels in a 3D scene are free space, such dense construction is inefficient for object detection and incurs significant computational overhead. To address this issue, we introduce a sparse volume construction strategy that constructs the 3D volume in a coarse-to-fine manner. As shown in Fig.~\ref{fig:method}(b), the key idea is to progressively upsample a coarse 3D volume, adaptively refining only the voxels that likely contain objects. Specifically, we first construct a coarse 3D volume $\mathbf{V}_{0} \in \mathbb{R}^{\frac{X}{2^{L}}\times \frac{Y}{2^{L}}\times \frac{Z}{2^{L}}\times C}$ with a spatial resolution of $(\frac{X}{2^{L}}, \frac{Y}{2^{L}}, \frac{Z}{2^{L}})$. This coarse volume captures the overall scene geometry, and can be used to identify regions that may contain objects for further refinement.

\textbf{Coarse-to-fine Refinement.} The overall refinement process composes of $L$ stages. As illustrated in Fig.~\ref{fig:method}(b), at the $l$-th stage, we first upsample the output volume of the $(l-1)$-th stage by a factor of 2, obtaining $\mathbf{V}_{l}^{\mathrm{init}} \in \mathbb{R}^{\frac{X}{2^{L-l}}\times \frac{Y}{2^{L-l}}\times \frac{Z}{2^{L-l}}\times C}$. Then, we estimate the occupancy probability of each voxel by
\begin{equation}
	\mathbf{O}_{l} = \mathcal{F}(\mathbf{V}_{l}^{\mathrm{init}}),
\end{equation}
where $\mathbf{O}_{l} \in \mathbb{R}^{\frac{X}{2^{L-l}}\times \frac{Y}{2^{L-l}}\times \frac{Z}{2^{L-l}}}$ denotes the occupancy probability, and $\mathcal{F}$ is a lightweight occupancy prediction head. Next, we select the positions with top-$k$ occupancy probability for feature refinement, formulated as
\begin{equation}
	\mathbf{V}_{l} = \mathbf{V}_{l}^{\mathrm{init}} + \mathcal{P}(\mathbf{P}_{l}, \{\mathbf{F}_{n}^\mathrm{2D} \}_{n=1}^{N}, \{\mathbf{D}_{n}\}_{n=1}^{N}),
\end{equation}
where $\mathbf{P}_{l}$ is the set of coordinates of the top $k\%$ points, and $\mathcal{P}$ denotes the geometry and context aware aggregation described in Sec.~\ref{subsec:gcap}. This strategy avoids redundant computation in free space, while effectively capturing fine structures in regions likely containing objects.

\begin{figure}[t]
	\centering
	\includegraphics[scale=0.4] {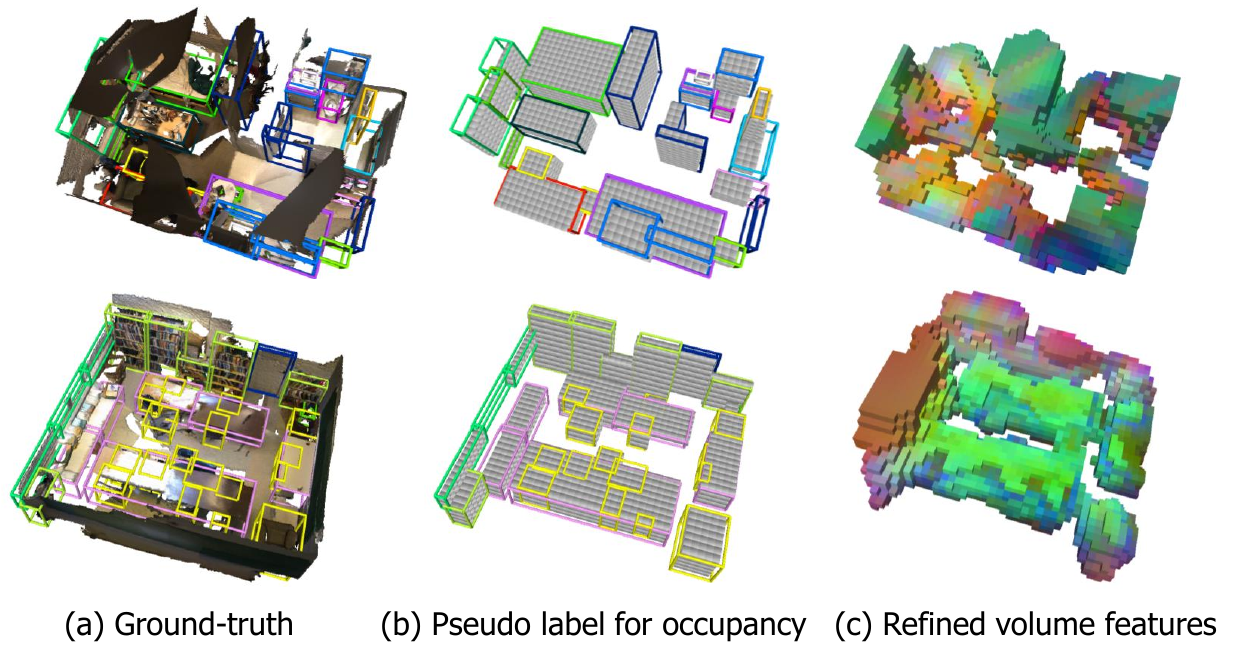}
	\vspace{-6mm}
	\caption{Visualization of our sparse volume construction. (a) Ground-truth 3D bounding boxes. (b) Pseudo-labels for occupancy supervision, generated from 3D bounding boxes. (c) Refined volume features. Our occupancy prediction network effectively filters free space, focusing the feature refinement on voxels that are likely to contain objects.}
	\label{fig:asvr_vis}
	\vspace{-4mm}
\end{figure}

\textbf{Supervision on Occupancy Probability.} A straightforward way is to supervise the occupancy probability via ground-truth scene geometry. However, it is not feasible when the precise geometry is unavailable~\cite{mvsdet_nips24, objectron_cvpr21}. To address this issue, we use the ground-truth 3D bounding boxes to generate pseudo labels for occupancy, providing a flexible supervision strategy. Given the 3D grid $\mathbf{G}_{l} \in \mathbb{R}^{\frac{X}{2^{L-l}}\times \frac{Y}{2^{L-l}}\times \frac{Z}{2^{L-l}}\times 3}$ at the $l$-th stage, the ground truth occupancy probability $\mathbf{O}_{l,\mathrm{gt}} \in \mathbb{R}^{\frac{X}{2^{L-l}}\times \frac{Y}{2^{L-l}}\times \frac{Z}{2^{L-l}}}$ is defined as:
\begin{equation}
	\mathbf{O}_{l,\mathrm{gt}}(x,y,z)= \begin{cases}
		1, \mathbf{G}_{l}(x,y,z) \text{ is inside any bounding box,}\\
		0, \text{ otherwise}.
	\end{cases}
\end{equation}
The network is supervised by the binary cross entropy loss between $\mathbf{O}_{l}$ and $\mathbf{O}_{l,\mathrm{gt}}$. Fig.~\ref{fig:asvr_vis} displays two examples of the pseudo-labels generated from 3D bounding boxes and refined volume features in the last refinement stage. It can be observed that our sparse volume construction effectively captures scene geometry and adaptively focuses on regions containing objects.

\subsection{Geometry and Context Aware Aggregation}
\label{subsec:gcap}

A detailed diagram of our geometry and context aware aggregation is shown in Fig.~\ref{fig:method}(c). To obtain features for a voxel with  center $\mathbf{p}=(x,y,z)^\top$, we first perform intra-view feature sampling to independently sample features from each view for initial information aggregation. Subsequently, we apply inter-view feature fusion to fuse the features from multiple views for further refinement.

\begin{figure}[t]
	\centering
	\includegraphics[scale=0.5] {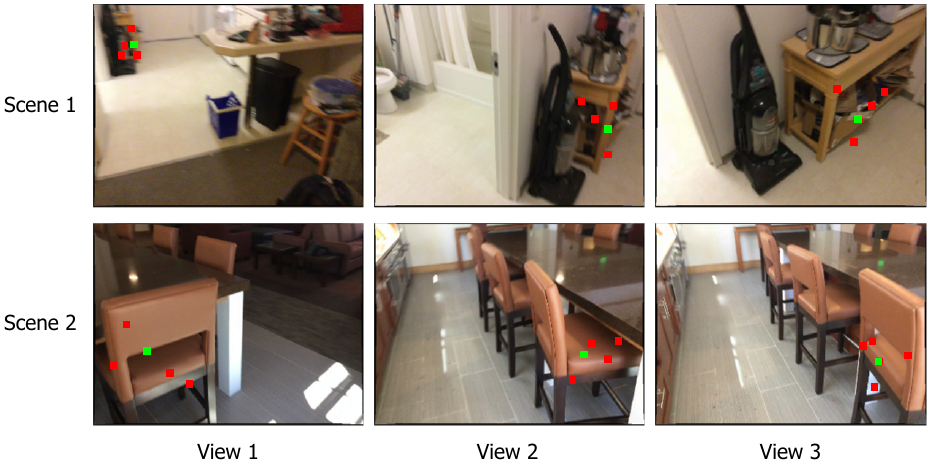}
	\vspace{-2mm}
	\caption{Visualization of sampling locations in our intra-view feature sampling. Green points represent positions derived from voxel centers and camera poses, while red points indicate deformable sampling locations. We note that the deformable attention is performed in the 3D pixel space, For clarity, the depth dimension is omitted in this visualization. }
	\vspace{-6mm}
	\label{fig:gcap_vis}
\end{figure}

\textbf{Intra-view Feature Sampling.} Previous methods~\cite{imvoxelnet_wacv22, imgeonet_iccv23, nerfdet_iccv23, cnrma_cvpr24, mvsdet_nips24} simply sample image features at the locations that derived from voxel centers and camera poses as the voxel features, limiting the receptive field of voxels. To address this problem, we introduce a 3D deformable attention mechanism~\cite{dfa3d_iccv23} to incorporate geometric and contextual information within an adaptive region. Specifically, for each view $n$, we lift the 2D image features to a 3D pixel space, formulated as 
\begin{equation}
	\mathbf{F}_{n}^\mathrm{3D} = \mathbf{F}_{n}^\mathrm{2D} \otimes \mathbf{D}_{n},
\end{equation}
where $\mathbf{F}_{n}^\mathrm{3D} \in \mathbb{R}^{H\times W\times D\times C}$ denotes the lifted 3D features, and $\otimes$ refers to the outer product conducted at the last dimension. Next, we project $\mathbf{p}$ to view $n$ as
\begin{equation}
	\mathbf{p}_{n} = (u_n, v_n, d_n)^\top = \mathbf{K}_{n} \mathbf{E}_{n} (\mathbf{p}, 1)^\top,
\end{equation}
where $\mathbf{p}_{n}$ is the coordinate in the 3D pixel space, $\mathbf{K}_{n}$ and $\mathbf{E}_{n}$ are the intrinsic and extrinsic matrices of view $n$, respectively. Instead of directly sampling $\mathbf{F}_{n}^\mathrm{3D}$ at $\mathbf{p}_{n}$ as the voxel features $\mathbf{V}_{n}(\mathbf{p})$, we take the sampled features as queries to aggregate information from neighboring regions, formulated as 
\begin{equation}
	\begin{aligned}
		\mathbf{V}_{n}(\mathbf{p}) &= \mathrm{DeformAttn}(\mathbf{p}_{n}, \phi(\mathbf{F}_{n}^\mathrm{3D}, \mathbf{p}_{n}), \mathbf{F}_{n}^\mathrm{3D})\\ 
		&= \sum_{m=1}^{M} A_{n,m} W \phi(\mathbf{F}_{n}^\mathrm{3D}, \mathbf{p}_{n}+\Delta \mathbf{p}_{n,m}),
	\end{aligned}
	\label{eq:dfa}
\end{equation}
where $M$ is the number of sampled points, $W$ is the matrix for value projection, and $\phi$ denotes the trilinear interpolation used to sample features from $\mathbf{F}_{n}^\mathrm{3D}$. $\Delta \mathbf{p}_{n,m}$ and $A_{n,m}$ are the 3D offset and attention weight of the $m$-th sampled point, respectively, and they are generated from the query $\phi(\mathbf{F}_{n}^\mathrm{3D}, \mathbf{p}_{n})$ via a linear layer. For simplicity, we exclude the multi-head operation in Eq.~\ref{eq:dfa}.  We show the locations of deformable sampling points across different views in Fig.~\ref{fig:gcap_vis}. Compared to the single point sampling, our geometry and context aware aggregation effectively integrates geometric and contextual information within a flexible region, thus enhancing the representation capabilities of voxel features.

\textbf{Inter-view Feature Fusion.} Due to variations in object appearance and size across different views, the features $\{\mathbf{V}_{n}(\mathbf{p})\}_{n=1}^{N}$ sampled from different views may differ significantly. To adaptively adjust the contribution of each view, we propose a multi-view attention mechanism. Specifically, we use the average pooling of features from all views $\mathbf{V}_\mathrm{avg}(\mathbf{p})$ as the query, while $\{\mathbf{V}_{n}(\mathbf{p})\}_{n=1}^{N}$ serves as both the key and value. This process can be formulated as
\begin{equation}
	\mathbf{V}(\mathbf{p}) = \mathrm{Attn}(\mathbf{V}_\mathrm{avg}(\mathbf{p}), \{\mathbf{V}_{n}(\mathbf{p})\}_{n=1}^{N}, \{\mathbf{V}_{n}(\mathbf{p})\}_{n=1}^{N}),
\end{equation}
where $\mathbf{V}(\mathbf{p})$ denotes the final features of the 3D point $\mathbf{p}$, and $\mathrm{Attn}(\cdot)$ refers to the standard attention operation~\cite{transformer_nips17, vit_iclr21}. Here, we assume that $\mathbf{p}$ can be projected to all views for notation simplicity. In practice, we discard any views where $\mathbf{p}$ is projected outside the image boundaries.

\textbf{Discussions.} Our geometry and context aware aggregation is highly inspired by DFA3D~\cite{dfa3d_iccv23}, upon which we introduce substantial modifications to better accommodate indoor scenes. DFA3D employs view-agnostic 3D queries to predict sampling offsets and weights across all views, which performs well in autonomous driving scenarios with fixed camera layouts. However, its effectiveness is limited in indoor environments, where camera poses vary significantly and objects exhibit large shape and scale differences across views. In contrast, our intra-view feature sampling leverages view-specific features as queries, enabling adaptive aggregation tailored to each individual view. Furthermore, our inter-view fusion module assigns learnable attention weights to each view's contribution, resulting in more consistent and robust scene-level voxel representations.

\begin{table*}
	\caption{Quantitative results and computational cost on the ScanNet dataset. * denotes the results are directly cited from~\cite{cnrma_cvpr24, mvsdet_nips24}.}
	\vspace{-2.5mm}
	\renewcommand{\tabcolsep}{8pt}{}
	\renewcommand\arraystretch{1.1}
	\centering
	\resizebox{\linewidth}{!}
	{
		\begin{tabular}{c|c|cc|cc|cc}
			\hline
			\multirow{2}{*}{Method} & \multirow{2}{*}{Voxel Resolution} & \multicolumn{2}{c|}{Performance} & \multicolumn{2}{c|}{Training Cost} & \multicolumn{2}{c}{Inference Cost} \\	
			\cline{3-8}
			& & mAP@0.25 & mAP@0.50 & Memory (GB) & Time (Hours) & Memory (GB) & FPS \\
			\hline
			\multicolumn{7}{l}{\textit{With ground-truth geometry supervision.}} \\
			\hline
			ImGeoNet*~\cite{imgeonet_iccv23} & 40$\times$40$\times$16 & 54.8 & 28.4 & 13 & 16 & 11 & 2.50 \\
			CN-RMA*~\cite{cnrma_cvpr24} & 256$\times$256$\times$96 & 58.6 & \textbf{36.8} & 43 & 242 & 12 & 0.26 \\
			\hline
			\multicolumn{7}{l}{\textit{Without ground-truth geometry supervision.}} \\
			\hline
			ImVoxelNet*~\cite{imvoxelnet_wacv22} & 40$\times$40$\times$16 & 46.7 & 23.4 & 11 & 13 & 9 & 2.60 \\
			NeRF-Det*~\cite{nerfdet_iccv23} & 40$\times$40$\times$16 & 53.5 & 27.4 & 13 & 14 & 12 & 1.30\\
			MVSDet*~\cite{mvsdet_nips24} & 40$\times$40$\times$16 & 56.2 & 31.3 & 35 & 36 & 28 & 0.87 \\
			SGCDet (Ours) & 40$\times$40$\times$16 & \textbf{61.2} & 35.2 & 20 & 19 & 14 & 1.46 \\
			\hline
		\end{tabular}
	}
	\vspace{-2.5mm}
	\label{tab:res_scannet}
\end{table*}

\subsection{DepthNet}
\label{subsec:depthnet}
\begin{figure}[t]
	\centering
	\includegraphics[scale=0.45] {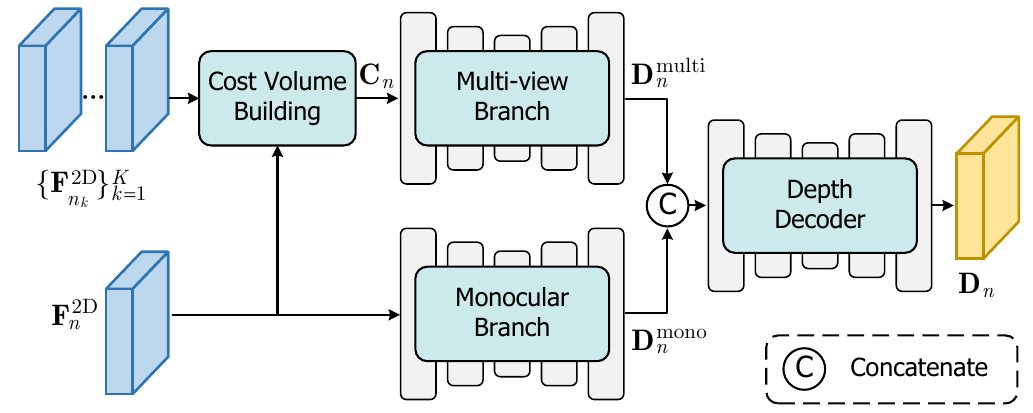}
	\vspace{-2mm}
	\caption{Detailed architecture of the DepthNet.}
	\vspace{-4mm}
	\label{fig:depthnet}
\end{figure}
The depth distributions provide geometric information for the 2D-to-3D projection process, whose accuracy significantly influences the final detection performance. To fully leverage multi-view images for accurate depth estimation, we introduce a simple yet effective DepthNet. As illustrated in Fig.~\ref{fig:depthnet}, it fuses both multi-view and monocular depth features for depth estimation, where the former provides geometric properties through feature matching, and the latter contributes detailed structures of input images.

For any view $n$ with 2D features $\mathbf{F}_{n}^\mathrm{2D}$, we select the nearest $K$ views with 2D features $\{\mathbf{F}_{n_k}^\mathrm{2D}\}_{k=1}^{K}$, and use plane sweeping~\cite{planesweep_cvpr96} to construct the cost volume. Specifically, we discretize the depth range $[d_\mathrm{min}, d_\mathrm{max}]$ into $D$ depth bins as $[d_1,...,d_i,...,d_D]$. For each depth plane $d_i$, we warp the 2D features of nearby views to view $n$ using camera matrices:
\begin{equation}
	\mathbf{F}_{n_k, d_i}^{2D} = \mathcal{W}(\mathbf{F}_{n_k}^{2D}, d_i, \mathbf{K}_{n}, \mathbf{E}_{n}, \mathbf{K}_{n_k}, \mathbf{E}_{n_k}),
\end{equation}
where $\mathcal{W}$ is warping operation in~\cite{mvsnet_eccv18, unify_tpami23}. Then, we build the cost volume $\mathbf{C}_{n} \in \mathbb{R}^{H\times W\times D}$ as:
\begin{equation}
	\mathbf{C}_{n}(h,w,i)=\frac{1}{K}\sum_{k=1}^{K}\frac{\mathbf{F}_{n}^{2D}(h,w) \cdot \mathbf{F}_{n_k, d_i}^{2D}(h,w)^\top}{\sqrt{C}}.
\end{equation}

We then process the cost volume and image features through two parallel branches, producing the multi-view depth features $\mathbf{D}_{n}^\mathrm{multi} \in \mathbb{R}^{H\times W\times D}$ and monocular depth features $\mathbf{D}_{n}^\mathrm{mono} \in \mathbb{R}^{H\times W\times C}$, respectively. Finally, these two features are concatenated and passed through a depth decoder to output the depth distributions $\mathbf{D}$.

\subsection{Overall Training Objective}
\label{subsec:losses}
The loss function of SGCDet comprises two components: detection loss $\mathcal{L}_\mathrm{det}$, and occupancy loss $\mathcal{L}_\mathrm{occ}$.

\textbf{Detection Loss.} Following~\cite{imvoxelnet_wacv22, imgeonet_iccv23, nerfdet_iccv23, mvsdet_nips24}, we use an anchor-free detection head. The detection loss $\mathcal{L}_\mathrm{det}$ consists of cross-entropy loss $\mathcal{L}_\mathrm{center}$ for centerness, IoU loss $\mathcal{L}_\mathrm{iou}$ for location, and focal loss $\mathcal{L}_\mathrm{cls}$ for classification, which can be formulated as $\mathcal{L}_\mathrm{det} = \mathcal{L}_\mathrm{center} + \mathcal{L}_\mathrm{iou} + \mathcal{L}_\mathrm{cls}.$

\textbf{Occupancy Loss.} We supervise $\mathbf{O}_{l}$ of each coarse-to-fine layer in the sparse volume construction as $\mathcal{L}_\mathrm{occ} = \sum_{l=1}^{L} \mathcal{L}_\mathrm{bce}(\mathbf{O}_{l}, \mathbf{O}_{l,gt})$, where $\mathcal{L}_\mathrm{bce}$ denotes the binary cross entropy loss.

The total loss is represented as
\begin{equation}
	\mathcal{L}=\mathcal{L}_\mathrm{det} + \lambda \mathcal{L}_\mathrm{occ},
\end{equation}
where we set the weight of occupancy loss $\lambda$ to $0.5$.

\section{Experiments}
\label{sec:exp}

\subsection{Datasets and Metrics}
We evaluate SGCDet on ScanNet~\cite{scannet_cvpr17}, ScanNet200~\cite{scannet200_eccv22}, and ARKitScenes~\cite{arkitscenes} datasets. ScanNet contains 1,201 scenes for training and 312 for testing, covering 18 categories. ScanNet200 extends ScanNet to 200 object categories with a broader range of object sizes. For both ScanNet and ScanNet200, we predict axis-aligned bounding boxes. ARKitScenes contains 4,498 scans for training and 549 for testing, with annotations for 17 classes. In this case, we detect oriented bounding boxes. We employ mean average precision (mAP) with thresholds of 0.25 and 0.5 for evaluation.

\subsection{Implementation Details}
\textbf{Network Details.} In alignment with~\cite{mvsdet_nips24}, we use 40 images for training and 100 images for testing. We employ ResNet-50~\cite{resnet_cvpr16} with a feature pyramid network (FPN)~\cite{fpn_cvpr17} as the image backbone. The spatial resolutions of input images and 2D feature maps are $320\times 240$ and $80\times 60$, respectively. For DepthNet, we set the depth range and depth bins to $[0.2m, 5m]$ and $12$, respectively, and use $2$ nearest views to construct the cost volume. The sparse volume construction has $2$ refinement stages, and we select the voxels with top $25\%$ occupancy probability for refinement. The number of sampling points in the deformable attention is set to $4$. 

We present two variants of our network, SGCDet and SGCDet-L, with channel dimensions of 256 and 128, respectively. SGCDet computes a 3D volume with a spatial resolution of $40 \times 40 \times 16$ and a voxel size of $0.2\,\text{m} \times 0.2\,\text{m} \times 0.16\,\text{m}$, while SGCDet-L produces a 3D volume with a higher spatial resolution of $80 \times 80 \times 32$ and a finer voxel size of $0.1\,\text{m} \times 0.1\,\text{m} \times 0.08\,\text{m}$.

\textbf{Training Setup.} We adopt the AdamW~\cite{AdamW} optimizer, and set the maximum learning rate to $0.0002$. The cosine decay strategy~\cite{cosdecay} is used to decrease the learning rate. The models are trained on NVIDIA A6000 GPUs. We train for 12 epochs on the ScanNet and ARKitScenes datasets, and for 30 epochs on the ScanNet200 datasets.

\subsection{Quantitative Results}

\begin{table}
	\caption{Quantitative results on the ScanNet200 dataset. * denotes the results are directly cited from~\cite{imgeonet_iccv23}. The voxel resolution of all approaches is 80$\times$80$\times$32.}
	\vspace{-2.5mm}
	\renewcommand\tabcolsep{15pt}{}
	\renewcommand\arraystretch{1.1}
	\centering
	\resizebox{\linewidth}{!}
	{
		\begin{tabular}{c|cccc}
			\hline
			\multirow{2}{*}{Method}  & \multicolumn{4}{c}{Performance (mAP@0.25)} \\
			\cline{2-5}
			&  Total & Head & Common & Tail \\
			\hline
			ImVoxelNet*~\cite{imvoxelnet_wacv22} & 19.0 & 34.1 & 14.0 & 7.7 \\
			ImGeoNet*~\cite{imgeonet_iccv23} & 22.3 & 38.1 & 17.3 & 9.7 \\
			SGCDet-L (Ours) & \textbf{28.9} & \textbf{46.0} & \textbf{24.0} & \textbf{14.9} \\
			\hline
		\end{tabular}
	}
	\vspace{-2mm}
	\label{tab:res_scannet200}
\end{table}

\begin{table}
	\caption{Quantitative results on the ARKitScenes dataset. * denotes the results are directly cited from~\cite{cnrma_cvpr24, mvsdet_nips24}.}
	\vspace{-2.5mm}
	\renewcommand{\tabcolsep}{6pt}{}
	\renewcommand\arraystretch{1.1}
	\centering
	\resizebox{\linewidth}{!}
	{
		\begin{tabular}{c|c|cc}
			\hline
			Method & Voxel Resolution & mAP@0.25 & mAP@0.50 \\
			\hline
			\multicolumn{3}{l}{\textit{With ground-truth geometry supervision.}} \\
			\hline
			ImGeoNet*~\cite{imgeonet_iccv23} & 40$\times$40$\times$16 & 60.2 & 43.4 \\
			CN-RMA*~\cite{cnrma_cvpr24} & 192$\times$192$\times$80 & 67.6 & 56.5 \\
			\hline
			\multicolumn{3}{l}{\textit{Without ground-truth geometry supervision.}} \\
			\hline
			ImVoxelNet*~\cite{imvoxelnet_wacv22} & 40$\times$40$\times$16 & 27.3 & 4.3 \\
			NeRF-Det*~\cite{nerfdet_iccv23} & 40$\times$40$\times$16 & 39.5 & 21.9 \\
			MVSDet*~\cite{mvsdet_nips24} & 40$\times$40$\times$16 & 42.9 & 27.0 \\
			ImVoxelNet~\cite{imvoxelnet_wacv22} & 40$\times$40$\times$16 & 58.0 & 33.2 \\
			NeRF-Det~\cite{nerfdet_iccv23} & 40$\times$40$\times$16 & 60.4 & 38.3 \\
			MVSDet~\cite{mvsdet_nips24} & 40$\times$40$\times$16 & 60.7 & 40.1 \\
			SGCDet (Ours) & 40$\times$40$\times$16 & 62.3 & 44.7 \\
			SGCDet-L (Ours) & 80$\times$80$\times$32 & \textbf{70.4} & \textbf{57.0} \\
			\hline
		\end{tabular}
	}
	\vspace{-2mm}
	\label{tab:res_arkit}
\end{table}

\begin{figure*}[t]
	\centering
	\includegraphics[scale=0.6] {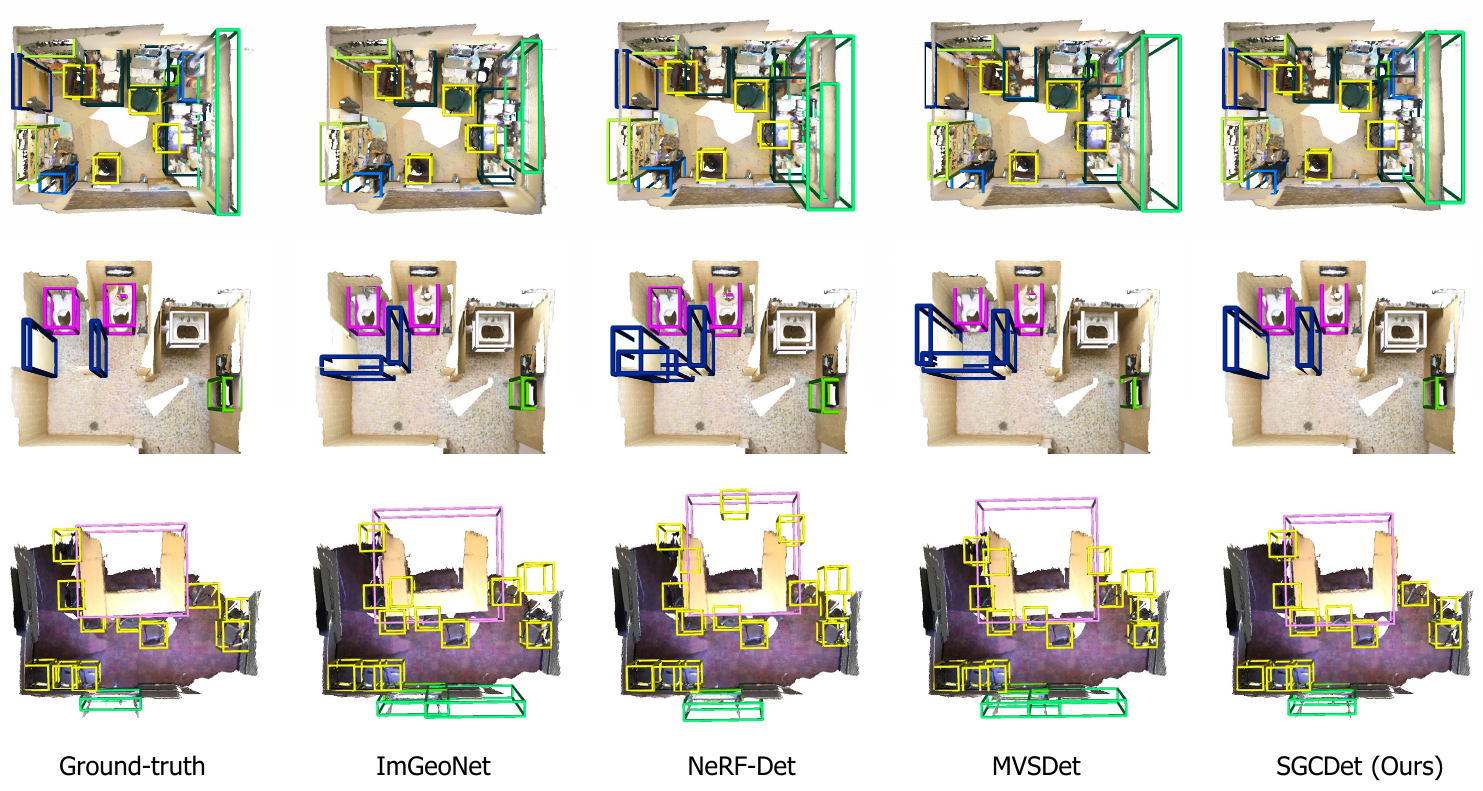}
	\vspace{-2.5mm}
	\caption{Qualitative comparison of different methods on the ScanNet dataset.}
	\label{fig:res_vis}
	\vspace{-2.5mm}
\end{figure*}

\begin{table}
	\caption{Ablation on the geometry and context aware aggregation. `2D Deform.' and `3D Deform.' denote the deformable attention is performed on 2D features $\mathbf{F}_{n}^\mathrm{2D}$ and lifted 3D features $\mathbf{F}_{n}^\mathrm{3D}$, respectively. `MV Attn.' denotes the multi-view attention.}
	\vspace{-2.5mm}
	\renewcommand{\tabcolsep}{6pt}{}
	\renewcommand\arraystretch{1.1}
	\centering
	\resizebox{\linewidth}{!}
	{
		\begin{tabular}{c|cc|c|cc}
			\hline
			Setting & 2D Deform. & 3D Deform. & MV Attn. & mAP@0.25 & mAP@0.50 \\
			\hline
			(a) &  &  &  & 56.0 & 29.8 \\
			(b) & \checkmark &  &  & 56.2 & 30.5 \\
			(c) &  & \checkmark &  & 59.5 & 34.1 \\
			\hline
			(d) &  & \checkmark & \checkmark & \textbf{61.2} & \textbf{35.2} \\
			\hline
		\end{tabular}
	}
	\label{tab:ablation_gcap}
	\vspace{-2mm}
\end{table}

\begin{table*}
	\caption{Ablation on the sparse volume reconstruction, including the number of refinement stages and the selection ratio for refinement. The setting (e) is used in our SGCDet.}
	\vspace{-2.5mm}
	\renewcommand{\tabcolsep}{8pt}{}
	\renewcommand\arraystretch{1.1}
	\centering
	\resizebox{\linewidth}{!}
	{
		\begin{tabular}{c|c|cc|cc|cc}
			\hline
			\multirow{2}{*}{Setting} & \multirow{2}{*}{Voxel Resolution (Selection Ratio)} & \multicolumn{2}{c|}{Performance} & \multicolumn{2}{c|}{Training Cost} & \multicolumn{2}{c}{Inference Cost} \\	
			\cline{3-8}
			& & mAP@0.25 & mAP@0.50 & Memory (GB) & Time (Hours) & Memory (GB) & FPS \\
			\hline
			(a) & 40$\times$40$\times$16 (100\%) & 61.0 & 36.0 & 31 & 24 & 22 & 1.33 \\
			\hline
			(b) & 20$\times$20$\times$8 (100\%) + 40$\times$40$\times$16 (25\%) & 60.6 & 35.6 & 21 & 21 & 14 & 1.40 \\
			\hline
			(c) & 10$\times$10$\times$4 (100\%) + 20$\times$20$\times$8 (100\%) + 40$\times$40$\times$16 (100\%) & 61.3 & 36.2 & 34 & 26 & 26 & 1.28 \\
			(d) & 10$\times$10$\times$4 (100\%) + 20$\times$20$\times$8 (50\%) + 40$\times$40$\times$16 (50\%) & 60.9 & 35.4 & 22 & 22 & 15 & 1.40 \\
			\rowcolor{gray!15} (e) & 10$\times$10$\times$4 (100\%) + 20$\times$20$\times$8 (25\%) + 40$\times$40$\times$16 (25\%) & 61.2 & 35.2 & 20 & 19 & 13 & 1.46 \\
			(f) & 10$\times$10$\times$4 (100\%) + 20$\times$20$\times$8 (10\%) + 40$\times$40$\times$16 (10\%) & 57.0 & 31.7 & 19 & 19 & 13 & 1.53 \\
			\hline
		\end{tabular}
	}
	\vspace{-2mm}
	\label{tab:ablation_asvr}
\end{table*}

We compare our method with the previous state-of-the-art approaches, including ImVoxelNet~\cite{imvoxelnet_wacv22}, ImGeoNet~\cite{imgeonet_iccv23}, NeRF-Det~\cite{nerfdet_iccv23}, CN-RMA~\cite{cnrma_cvpr24}, and MVSDet~\cite{mvsdet_nips24}. It is noted that ImGeoNet and CN-RMA require ground-truth geometry for training. Additionally, CN-RMA relies on a time-consuming multi-stage training pipeline, and uses a higher voxel resolution compared to other approaches.

Table~\ref{tab:res_scannet} lists the performance and computational cost on the ScanNet dataset. The computational cost is measured on a single NVIDIA A6000 GPU. SGCDet achieves an mAP@0.25 of 61.2 and an mAP@0.50 of 35.2, surpassing all comparison approaches without using ground-truth geometry for supervision. Compared to the previous state-of-the-art approach MVSDet, SGCDet attains gains of 5.0 and 3.9 in terms of mAP@0.25 and mAP@0.50, respectively. Furthermore, SGCDet even achieves better or comparable performance than those approaches requiring ground-truth geometry during training. In terms of computational cost, SGCDet substantially reduces both training and inference costs compared to CN-RMA and MVSDet, which explicitly estimate geometry for feature lifting. Although ImVoxelNet, ImGeoNet, and NeRF-Det are efficient, their detection performance is notably lower than ours. Overall, SGCDet achieves a remarkable balance between accuracy and computational cost, while eliminating reliance on ground-truth geometry. We further evaluate SGCDet-L on the ScanNet200 dataset, with the results shown in Table~\ref{tab:res_scannet200}. The object sizes decrease from the head to the tail group. SGCDet-L consistently outperforms other approaches, demonstrating strong robustness to small objects and complex scenes with dense object distributions.

Table~\ref{tab:res_arkit} presents the results on the ARKitScenes dataset. It is observed that some approaches exhibit a substantial performance drop compared to their results on ScanNet. This discrepancy arises because the coordinate origin of 3D scenes in ARKitScenes is positioned far from the scene center, causing the perception region of the constructed 3D volume to fail to cover the 3D scenes. To ensure a fair comparison, we follow ImGeoNet~\cite{imgeonet_iccv23} to relocate the coordinate origin to the center of the input camera poses and reproduce these approaches. As shown in Table~\ref{tab:res_arkit}, SGCDet consistently provides the best performance compared to all approaches with the same 3D voxel resolution. Moreover, our SGCDet-L, with a voxel resolution of $80\times80\times32$, outperforms CN-RMA, which uses a higher voxel resolution of $192\times192\times80$ and ground-truth geometry supervision. These results further demonstrate the effectiveness of our proposed SGCDet.

\subsection{Qualitative Results}
Fig.~\ref{fig:res_vis} presents visualizations of predicted 3D bounding boxes obtained from ImGeoNet\cite{imvoxelnet_wacv22}, NeRF-Det~\cite{nerfdet_iccv23}, MVSDet~\cite{mvsdet_nips24}, and our proposed SGCDet. It is observed that the comparison approaches often miss some objects or predict incorrect bounding boxes in free space. In contrast, SGCDet produces more accurate detection results.

\subsection{Ablation Studies}
We conduct ablation studies on the ScanNet dataset.

\textbf{Ablation on the geometry and context aware aggregation.} Table~\ref{tab:ablation_gcap} shows the ablation study of the geometry and context aware aggregation. Setting (a) serves as our baseline, employing a single-point sampling strategy for feature lifting. Settings (b) and (c) examine the impact of aggregating image features within a deformable region. Although 2D deformable attention enlarges the receptive field of voxels, it suffers from depth ambiguity, resulting in limited performance gains. In contrast, our 3D deformable attention simultaneously incorporates geometric and contextual information within an adaptive region, leading to notable improvements of 3.3 and 3.6 in mAP@0.25 and mAP@0.50, respectively. The performance is further enhanced by integrating multi-view attention, which dynamically adjusts contributions from different views (setting (d)).

\textbf{Ablation on the sparse volume reconstruction.} Table~\ref{tab:ablation_asvr} presents a detailed analysis of the sparse volume reconstruction. Setting (a) is the baseline that directly builds the 3D volume with a fixed resolution of $40 \times 40 \times 16$. Comparing setting (a), (b), and (e), we observe that the coarse-to-fine strategy significantly reduces computational cost, while maintaining performance. We further vary the selection ratio for refinement in settings (c)-(f).  Although reducing the selection ratio improves efficiency, an overly small selection ratio (\eg, $10\%$) may miss object regions, degrading detection accuracy. To balance both accuracy and computational overhead, we set the selection ratio to $25\%$.

\textbf{Ablation on the occupancy loss.} As shown in Table~\ref{tab:ablation_occloss}, removing occupancy loss leads to a performance drop of 6.7 mAP@0.25 and 6.2 mAP@0.50, demonstrating the importance of explicit occupancy supervision. Thanks to our pseudo-labeling strategy based on 3D bounding boxes, we eliminate the reliance on ground-truth scene geometry.

\begin{table}
	\caption{Ablation on the occupancy loss.}
	\vspace{-2.5mm}
	\renewcommand{\tabcolsep}{15pt}{}
	\renewcommand\arraystretch{1.1}
	\centering
	\resizebox{\linewidth}{!}
	{ 
		\begin{tabular}{c|cc}
			\hline
			Setting & mAP@0.25 & mAP@0.50 \\
			\hline
			w/o occupancy loss & 54.5 & 29.0 \\
			w/ occupancy loss & \textbf{61.2} & \textbf{35.2} \\
			\hline
		\end{tabular}
	}
	\vspace{-2.5mm}
	\label{tab:ablation_occloss}
\end{table}

\textbf{A deeper look at the pseudo-labeling strategy.} The 3D bounding boxes may produce noisy occupancy labels, particularly at the box boundaries or in cluttered scenes. However, these noisy occupancy labels are only used in the training stage, serving as an explicit supervision for occupancy prediction. For inference, the top 25\% selection for refinement ensures sufficient coverage of occupied regions, including areas not annotated by pseudo-labels (Fig.~\ref{fig:asvr_vis}). To further assess the impact of noisy bounding box annotations, we randomly drop 15\% of the ground-truth boxes, and apply 15\% random scaling to the remaining boxes during training. These imperfect annotations affect both occupancy prediction and the learning of 3D detection. Nevertheless, Table~\ref{tab:ablation_label} demonstrates that our SGCDet exhibits higher robustness compared to ImGeoNet~\cite{imgeonet_iccv23}, which uses ground-truth geometry supervision.

\begin{table}
	\caption{Ablation on the 3D bounding boxes label quality.}
	\vspace{-2.5mm}
	\renewcommand\arraystretch{1.1}
	\renewcommand\tabcolsep{5pt}{}
	\centering
	\resizebox{\linewidth}{!}
	{
		\begin{tabular}{c|cc|cc}
			\hline
			\multirow{2}{*}{Label quality} & \multicolumn{2}{c|}{Ground-truth labels} & \multicolumn{2}{c}{Noisy and incomplete labels} \\
			\cline{2-5}
			& mAP@0.25 & mAP@0.50 & mAP@0.25 & mAP@0.50 \\
			\hline
			ImGeoNet~\cite{imgeonet_iccv23} & 54.8 & 28.4 & 54.0 ($\downarrow$ 0.8) & 26.2 ($\downarrow$ 2.2) \\
			SGCDet (Ours)  & \textbf{61.2} & \textbf{35.2} & \textbf{60.7} ($\downarrow$ 0.5) & \textbf{33.6} ($\downarrow$ 1.6) \\
			\hline
		\end{tabular}
	}
	\vspace{-2.5mm}
	\label{tab:ablation_label}
\end{table}

\textbf{Ablation on the DepthNet.} We present the ablation analysis of the DepthNet in Table~\ref{tab:ablation_depthnet}. As shown in setting (a)-(c), removing any component of the DepthNet leads to a decrease of the detection accuracy. We then evaluate the influence of the depth quality. Adding depth supervision (setting (d)) achieves an mAP@0.25 of 62.2 and an mAP@0.50 of 37.1. Notably, these results even surpass CN-RMA~\cite{cnrma_cvpr24} that needs a multi-stage training pipeline and ground-truth scene geometry for supervision. Setting (e) refers to directly using the ground-truth depth as input, indicating the upper bound of our model. It reveals a big improvement space by further study on more accurate depth estimation.

\begin{table}
	\caption{Ablation on modules in DepthNet and depth quality.}
	\vspace{-2.5mm}
	\renewcommand{\tabcolsep}{15pt}{}
	\renewcommand\arraystretch{1.1}
	\centering
	\resizebox{\linewidth}{!}
	{
		\begin{tabular}{l|cc}
			\hline
			Setting & mAP@0.25 & mAP@0.50 \\
			\hline
			(a) w/o monocular branch & 59.6 & 33.8 \\
			(b) w/o multi-view branch & 57.7 & 32.1 \\
			\hline
			(c) full model & 61.2 & 35.2 \\
			\hline
			(d) w/ depth supervision & 62.2 & 37.1 \\
			(e) ground-truth depth & 64.3 & 42.3 \\
			\hline
		\end{tabular}
	}
	\vspace{-2.5mm}
	\label{tab:ablation_depthnet}
\end{table}

\section{Conclusions}
\label{sec:con}
We have proposed SGCDet, a novel multi-view indoor 3D object detection framework. To enhance the representation capability of voxel features, we introduce a geometry and context aware aggregation module that adaptively integrates image features across multiple views. Additionally, we develop a sparse volume construction strategy that selectively refines voxels with high occupancy probability, significantly reducing redundant computation in free space. Our framework is trained using only 3D bounding boxes for supervision, eliminating the need for ground-truth scene geometry. Experimental results demonstrate that SGCDet achieves state-of-the-art performance on the ScanNet, ScanNet200, and ARKitScenes datasets.

\section*{Acknowledgments}
This work was supported in part by the National Key Research and Development Program of China under grant 2023YFB3209800, in part by the National Natural Science Foundation of China under grant 62301484, in part by the Ningbo Natural Science Foundation of China under grant 2024J454, and in part by the Aeronautical Science Foundation of China under grant 2024M071076001. We also thank the generous help from Sijin Li, Zhejiang University.

{
	\small
	\bibliographystyle{ieeenat_fullname}
	\bibliography{main}
}


\end{document}